\documentclass{llncs}
\usepackage{graphicx}
\usepackage{hyperref}

\begin{document}

\pagestyle{headings}

\mainmatter

\title{Evolving reversible circuits for \\ the even-parity problem}

\author{Mihai Oltean}
\institute{Department of Computer Science\\
Faculty of Mathematics and Computer Science\\
Babe\c s-Bolyai University, Kog\u alniceanu 1\\
Cluj-Napoca, 3400, Romania.
\\
\email{mihai.oltean@gmail.com}\\
\url{https://mihaioltean.github.io}
}
\maketitle

\begin{abstract} 

Reversible computing basically means computation with less or not at all electrical power. Since the standard binary gates are not usually reversible we use the Fredkin gate in order to achieve reversibility. An algorithm for designing reversible digital circuits is described in this paper. The algorithm is based on Multi Expression Programming (MEP), a Genetic Programming variant with a linear representation of individuals. The case of digital circuits for the even-parity problem is investigated. Numerical experiments show that the MEP-based algorithm is able to easily design reversible digital circuits for up to the even-8-parity problem.

\end{abstract}

\section{Introduction}

The ultimate purpose of reversible computing is to perform computations less or not at all electrical power. Logically reversible operations occupy a central role in considerations of the fundamental physical limits of information handling \cite{landauer}. The early work of Landauer showed that energy dissipation occurs during the destruction of information of the previous state of the system rather than the acquisition of information during the computational process. Subsequently, Bennett showed that computation could be carried out completely with operations that are logically reversible, i.e., operations in which the output uniquely defines the input \cite{bennett}. 

One such reversible logic element is the Fredkin gate (FG) \cite{fredkin,langdon} which contains 3 inputs and 3 outputs. Fredkin gate constitute a complete set of operators in that any logic operation (e.g., AND, OR, NOT) can be constructed from a combination of FGs.

In this paper, we propose a variant of the Multi Expression Programming (MEP) \cite{oltean_eh,oltean_parity} for designing reversible digital circuits for the even-parity problem. We choose to apply the MEP-based technique to the even-parity problems because according to Koza \cite{koza2} these problems appear to be the most difficult Boolean functions to be detected via a blind random search.

Standard GP was able to solve up to even-5 parity when the set of gates $F$=\{AND, OR, NAND, NOR\} is used \cite{koza1}. Improvements, such as Automatically Defined Functions \cite{koza2} and Sub-symbolic node representation \cite{poli2}, allows GP programs to solve larger instances of the even-parity problem. Using MEP and reversible gates we are able to evolve a solution up to even-8-parity function using a reasonable population size.

The paper is organized as follows. MEP technique is briefly described in section \ref{mep}. The way in which MEP can be applied for reversible circuits is introduced in section \ref{ref_rev}. Several numerical experiments for designing reversible digital circuits are performed in section \ref{exp}. A comparison with standard digital circuits is described in section \ref{comparison}. Further research directions are indicated in section \ref{further}.

\section{Basic on MEP}\label{mep}

The \textit{Multi Expression Programming} (MEP) \cite{oltean_parity_fea,oltean_eh,oltean_parity} technique is briefly described in this section.

\subsection{Individual Representation}

MEP genes are represented by substrings of a variable length. The number of 
genes per chromosome is constant and it defines the length of the 
chromosome. Each gene encodes a terminal or a function symbol. A gene 
encoding a function includes references towards the function arguments. 
Function arguments always have indices of lower values than the position of 
that function in the chromosome.

This representation is similar to the way in which \textbf{\textit{C}} and 
\textbf{\textit{Pascal}} compilers translate mathematical expressions into 
machine code.

MEP representation ensures that no cycle arises while the 
chromosome is decoded (phenotypically transcripted). According to the 
representation scheme the first symbol of the chromosome must be a 
terminal symbol. In this way only syntactically correct programs (MEP 
individuals) are obtained.\\

\textbf{Example}\\

We employ a representation where the numbers on the left positions stand for 
gene labels (or memory addresses). Labels do not belong to the chromosome, 
they are provided here only for explanation purposes.

For this example, we use the set of functions $F$ = {\{}+, *{\}} and the set of 
terminals $T$ = {\{}$a$, $b$, $c$, $d${\}}. An example of chromosome using the sets $F$ and 
$T$ is given below:\\

1: $a$

2: $b$

3: + 1, 2

4: $c$

5: $d$

6: + 4, 5

7: * 3, 6\\

\subsection{Decoding MEP Chromosome and Fitness Assignment Process}

In this section we described the way in which MEP individuals are 
translated into computer programs and the way in which the fitness of these 
programs is computed.

This translation is achieved by reading the chromosome top-down. A terminal 
symbol specifies a simple expression. A function symbol specifies a complex 
expression obtained by connecting the operands specified by the argument 
positions with the current function symbol.

For instance, genes 1, 2, 4 and 5 in the previous example encode simple 
expressions formed by a single terminal symbol. These expressions are:\\

$E_{1}=a$,

$E_{2}=b$,

$E_{4}=c$,

$E_{5}=d$,\\

Gene 3 indicates the operation + on the operands located at positions 1 and 
2 of the chromosome. Therefore gene 3 encodes the expression:\\

$E_{3}=a+b$.\\

Gene 6 indicates the operation + on the operands located at positions 4 and 
5. Therefore gene 6 encodes the expression:\\

$E_{6}=c+d$.\\

Gene 7 indicates the operation * on the operands located at position 3 and 
6. Therefore gene 7 encodes the expression:\\

$E_{7}=(a+b)*(c+d)$.\\

\noindent
$E_{7}$ is the expression encoded by the whole chromosome.

There is neither practical nor theoretical evidence that one of these 
expressions is better than the others. Moreover Wolpert and McReady \cite{wolpert1} 
proved that we cannot use the search algorithm's behavior so far for a 
particular test function to predict its future behavior on that function. Thus we cannot choose one of the expressions (let us say expression $E_{7}$) to store the output of the chromosome. Even this expression proves to be useful for the first 10 generations we cannot guarantee that it will be the best option for all generations.

This is why each MEP chromosome is allowed to encode a number of expressions 
equal to the chromosome length. Each of these expressions is considered as 
being a potential solution of the problem. 

This is very important because we can get many solutions within the same running time as in the case of one solution/chromosome.

The value of these expressions may be computed by reading the chromosome top 
down. Partial results are computed by Dynamic Programming \cite{bellman} and are stored 
in a conventional manner.

As MEP chromosome encodes more than one problem solution, it is interesting 
to see how the fitness is assigned. Usually the chromosome fitness is defined as the fitness of the best 
expression encoded by that chromosome. For instance, if we want to solve symbolic regression problems the fitness 
of each sub-expression $E_{i}$ may be computed using the formula:

\[
f(E_i ) = \sum\limits_{k = 1}^n {\left| {o_{k,i} - w_k } \right|} ,
\]

\noindent
where $o_{k,i}$ is the obtained result by the expression $E_{i}$ for the 
fitness case $k$ and $w_{k}$ is the targeted result for the fitness case $k$. In 
this case the fitness needs to be minimized.

The fitness of an individual is set to be equal to the lowest fitness of the 
expressions encoded in chromosome:

\[
f(C) = \mathop {\min }\limits_i f(E_i ).
\]

When we have to deal with other problems we compute the fitness of each 
sub-expression encoded in the MEP chromosome and the fitness of the entire 
individual is given by the fitness of the best expression encoded in that 
chromosome.

\subsection{Genetic Operators}

Search operators used within MEP algorithm are crossover and mutation. 
These operators preserve the chromosome structure. All offspring 
are syntactically correct expressions.

\subsubsection{Crossover}

By crossover two parents are selected and recombined. For instance, 
within the uniform recombination the offspring genes are taken randomly from 
one parent or another.\\

\textbf{Example}\\

Let us consider the two parents $Parent_{1}$ and $Parent_{2}$ given in Table \ref{table:16-1}. The two 
offspring $Offspring_{1}$ and $Offspring_{2}$ are obtained by uniform recombination as shown in Table \ref{table:16-1}.

\begin{table}[ht]
\caption{MEP uniform recombination.}
\label{table:16-1}
\begin{center}
\begin{tabular}
{p{60pt}p{60pt}p{60pt}p{60pt}}
\hline
\multicolumn{2}{p{104pt}}{Parents } & 
\multicolumn{2}{p{98pt}}{Offspring}  \\
$Parent_{1}$& 
$Parent_{2}$& 
$Offspring_{1}$& 
$Offspring_{2}$ \\
\hline
1: \textbf{\textit{b}} \par 2: \textbf{* 1, 1} \par 3: \textbf{+ 2, 1} \par 4: \textbf{\textit{a}} \par 5: \textbf{* 3, 2} \par 6: \textbf{\textit{a}} \par 7: \textbf{- 1, 4}& 
1: $a$ \par 2: $b$ \par 3: + 1, 2 \par 4: $c$ \par 5: $d$ \par 6: + 4, 5 \par 7: * 3, 6& 
1: $a$ \par 2: \textbf{* 1, 1} \par 3: \textbf{+ 2, 1} \par 4: $c$ \par 5: \textbf{* 3, 2} \par 6: + 4, 5 \par 7: \textbf{- 1, 4}& 
1: \textbf{\textit{b}} \par 2: $b$ \par 3: + 1, 2 \par 4: \textbf{\textit{a}} \par 5: $d$ \par 6: \textbf{\textit{a}} \par 7: * 3, 6 \\
\hline
\end{tabular}
\end{center}
\end{table}

\subsection{Mutation}

Each symbol (terminal, function or function pointer) in the chromosome may 
be the target of mutation operator. By mutation some symbols in the chromosome 
are changed with a fixed mutation probability $p_m$. To preserve the consistency of the chromosome its first gene 
must encode a terminal symbol.

\subsection{MEP Algorithm}

Standard MEP algorithm uses steady state \cite{syswerda1} as its underlying mechanism. MEP 
algorithm starts by creating a random population of individuals. The 
following steps are repeated until a given number of generations \footnote{In a steady-state algorithm, a generation is considered when the number of newly created individuals is equal to the population size.} is reached. 
Two parents are selected using a selection procedure. The parents are 
recombined in order to obtain two offspring. The offspring are considered 
for mutation. The best offspring replaces the worst individual in the 
current population if the offspring is better than the worst individual. 

The algorithm returns as its answer the best expression evolved along a 
fixed number of generations.

\section{Reversible computing}

The ultimate purpose of reversible computing is to perform computations less or not at all electrical power. Logically reversible operations occupy a central role in considerations of the fundamental physical limits of information handling \cite{landauer}. The early work of Landauer showed that energy dissipation occurs during the destruction of information of the previous state of the system rather than the acquisition of information during the computational process. Subsequently, Bennett showed that computation could be carried out completely with operations that are logically reversible, i.e., operations in which the output uniquely defines the input \cite{bennett}. 

One such reversible logic element is the Fredkin gate (FG) \cite{fredkin} which contains an input control channel A, and two additional input channels, B and C, which exchange values if A is set at 1 or will go through the gate unchanged if A is set at 0. Fredkin gates constitute a complete set of operators in that any logic operation (e.g., AND, OR, NOT) can be constructed from a combination of FGs \cite{fredkin}.

\begin{figure}[htbp]
\centerline{\includegraphics[width=4.75in,height=1.28in]{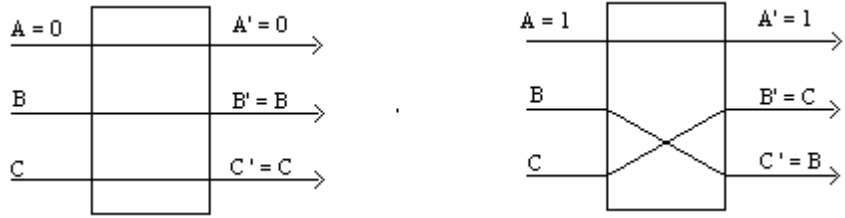}}
\caption{Fredkin gate has 3 inputs and 3 outputs. If A = 0 the outputs are identical with the inputs. If A = 1 the inputs B and C are swapped. We can easily reconstruct the input from the output.}
\label{fig_Fredkin}
\end{figure}

The Fredkin gate is depicted in Figure \ref{fig_Fredkin}.

\subsection{MEP for reversible circuits}\label{ref_rev}

The interpretation for a MEP chromosome needs to be modified because reversible gates have more than one output. Thus an MEP chromosome containing $N$ Fredkin gates actually provides $3*N$ outputs (plus the outputs provided directly from the inputs). MEP representation will be unchanged, but during the fitness evaluation we will have to handle more circuits than the case of standard gates. 

Another modification is related to the number of inputs. Two constant inputs 0 (always-OFF) and 1 (always-ON) have been added. These 2 inputs are very important in simulating the standard gates (such as NOT, AND) \cite{fredkin}. Moreover, without these 2 inputs we are not able to build a circuit for the even-parity problems. For instance, in the case of even-3-parity problem our circuits must signal 0 when all inputs are 1. But, the Fredkin gate can never generate a 0 value when all inputs are 1 (see Figure \ref{fig_Fredkin}).

\section{Numerical experiments}\label{exp}

Several numerical experiments for evolving reversible digital circuits are performed in this section.

\subsection{Test problem}

Our aim is to find a Boolean function that satisfies a set of fitness cases. The particular function that we want to find is the Boolean even-parity function. This function has $k$ Boolean arguments and it returns \textbf{T} 
(\textbf{True}) if an even number of its arguments are \textbf{T}. Otherwise 
the even-parity function returns \textbf{F} (\textbf{False}) \cite{koza2}. 
According to \cite{koza2} the Boolean even-parity functions appear to be the most 
difficult Boolean functions to detect via a blind random search.

The terminal set $T$ consists of the $k+2$ Boolean arguments $d_{0}$, $d_{1}$, $d_{2}$, ... $d_{k - 1}$, 0, 1.

The function set $F$ consists of one three-argument gate: the Fredkin gate. 

The set of fitness cases for this problem consists of the 2$^{k}$ combinations 
of the $k$ Boolean arguments. We have also added two constants inputs which are always signals 0 (respectively 1). These 2 fixed inputs are very important in simulating standard gates (such as NOT, AND, see \cite{fredkin} for more details). Thus each fitness case will have $k+2$ inputs and one output.

\subsection{Results}

In this section we perform several experiments with MEP for solving several instances of the even-parity problem. General parameter settings for MEP are given in Table \ref{tab_param}.

\begin{table}[ht]
\begin{center}
\caption{General parameters of the MEP algorithm for designing reversible circuits for the even-parity problem.}
\label{tab_param}
\begin{tabular}
{p{100pt}p{180pt}}
\hline
\textbf{Parameter}& 
\textbf{Value} \\
\hline
Mutation probability& 
0.2 \\
Crossover type& 
Uniform \\
Crossover probability& 
0.9 \\
Selection& 
$q$-tournament ($q$ = 1{\%} of the population size) \\
Function set& 
$F$ = {\{}Fredkin gate{\}} \\
\hline
\end{tabular}
\end{center}
\end{table}

For reducing the chromosome length we keep all the terminals on the first positions of the MEP chromosomes. 

The results along with the particular parameters used for obtaining them are given in Table \ref{tab_results}. Success rate is computed as the number of successful runs over the total number of runs.

\begin{table}[ht]
\begin{center}
\caption{Success rate of the MEP-based algorithm for evolving reversible digital circuits. Success rate is computed over 100 independent runs. Circuit size is the minimum number of gates obtained in one of the successfull runs.}
\label{tab_results}
\begin{tabular}
{p{60pt}p{50pt}p{70pt}p{70pt}p{40pt}p{40pt}}
\hline
\textbf{Problem}&
\textbf{Pop size}&
\textbf{Number of generations}&
\textbf{Chromosome length}&
\textbf{Success rate \%}&
\textbf{Circuit size}
\\
\hline
even-3-parity &
1000&
50&
10&
95&
3 \\
even-4-parity &
1000&
50&
15&
35&
4 \\
even-5-parity&
1000&
100&
20&
15&
5\\
even-6-parity&
2000&
200&
30&
18&
6 \\
even-7-parity& 
3000&
500&
30&
29&
8 \\
even-8-parity& 
5000&
500&
30&
11&
12 \\
\hline
\end{tabular}
\end{center}
\end{table}

\begin{figure}[htbp]
\centerline{\includegraphics[width=4in,height=2in]{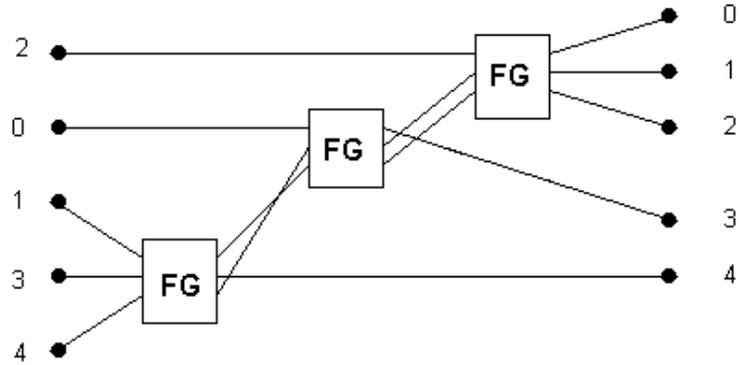}}
\caption{The shortest evolved reversible digital circuit for the even-3-parity problem. Input 3 always signals 0 and input 4 always signals 1. Output 1 provides the result for the even-parity problem. The other outputs are used only for achieving the reversibility. FG stands for the Fredkin gate.}
\label{even3}
\end{figure}

\begin{figure}[htbp]
\centerline{\includegraphics[width=5in,height=2.19in]{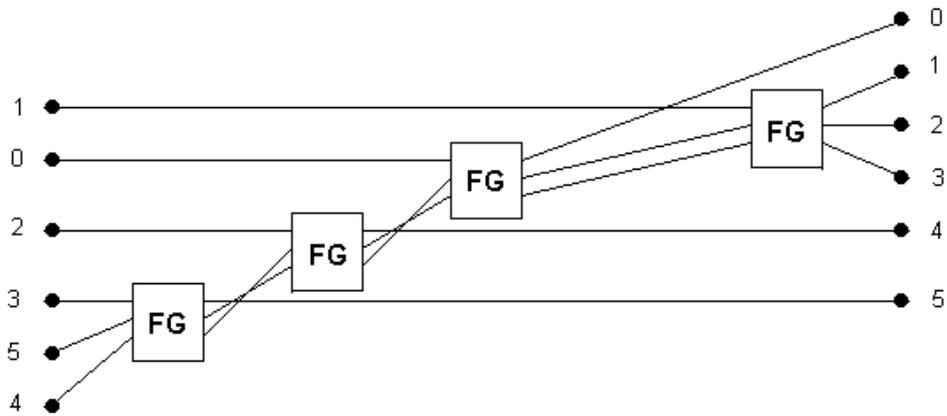}}
\caption{The shortest evolved evolved reversible digital circuit for the even-4-parity problem. Output 3 provides the result for the even-parity problem. The other outputs are used only for achieving the reversibility. FG stands for the Fredkin gate.}
\label{even4}
\end{figure}

Table \ref{tab_results} shows that MEP algorithm is able to evolve reversible circuits for the even-parity problem. The shortest (regarding the number of gates) evolved reversible circuits for the even-3-parity and even-4-parity problem are depicted in Figures \ref{even3} and \ref{even4}.

\subsection{Comparison with standard approaches}\label{comparison}

Multi Expression Programming has been used \cite{oltean_parity} for designing standard digital circuits for the even-parity problem. Using the gates AND, OR, NAND, NOR we have been able to evolve up to even-5-parity problem using a population of 4000 individuals with 600 genes each evolved for 50 generations. The shortest evolved standard digital circuit has 6 gates for the even-3-parity problem, and 9 gates for the even-4-parity problem, whereas the reversible ones requires 4 (even-3-parity) and 5 (even-4-parity) gates. The first remark is that reversible circuits might require less gates than the standard circuits.

However, when the entire set of 16 binary gates (including EQ, NOT, etc) was employed \cite{oltean_parity} the length of the evolved standard circuit is considerable shorter. Only 4 gates are required for a circuit implementing the even-5-parity problem and 5 standard gates are required for the even-6-parity problem \cite{oltean_parity}. The results obtained by using the Fredkin gate are similar (regarding the number of gates) to those obtained using the entire set of 16 gates with 2 binary inputs.

\section{Conclusions and further work}\label{further}

An algorithm based on Multi Expression Programming has been used for designing reversible digital circuits. Numerical experiments have shown the ability of this algorithm to design reversible digital circuits. When compare to the standard circuits, we can see that the number of outputs of the reversible ones is larger than the case of the standard circuits. This is in full concordance with other studies \cite{fredkin} which have shown that reversible computing requires addition storage space. Further experiments will try to minimize the number of outputs required by the reversible digital circuit. However, this number cannot be less than 3 (the number of outputs of the Fredkin gate).

We will also be interested in extracting general principles from the evolved circuits in order to quickly build larger size reversible circuits. For instance, Cartesian Genetic Programming was used \cite{miller1} for discovering of ripple-carry adder which is widely used for building large scale multipliers and adders. The evolution of Automatically Defined Functions \cite{koza2} will also be an interesting aspect for reversible digital circuits.

The method will be used for designing other interesting digital circuits such as reversible adders and multipliers. Other reversible gates, such as CCNOT, will be considered in further experiments \cite{langdon}.

\end{document}